\documentclass{article}

 \usepackage[dblblindworkshop, final]{neurips_2025}
\workshoptitle{Foundation Models for the Brain and Body}

\usepackage[utf8]{inputenc} 
\usepackage[T1]{fontenc}    
\usepackage{hyperref}       
\usepackage{url}            
\usepackage{booktabs}       
\usepackage{amsfonts}       
\usepackage{nicefrac}       
\usepackage{microtype}      
\usepackage{xcolor}         
\usepackage{graphicx}
\usepackage{cleveref}

\title{Learning Time-Scale Invariant Population-Level Neural Representations}

\author{
    Eshani Patel\textsuperscript{1}~~~~Yisong Yue\textsuperscript{1}~~~~Geeling Chau\textsuperscript{2}
    \\
  Computing \& Mathematical Sciences\textsuperscript{1},  Computation \& Neural Systems\textsuperscript{2}\\
  California Institute of Technology
  , Pasadena, CA 91125 \\
  \texttt{ejpatel@alumni.caltech.edu} ~ \texttt{\{yyue, gchau\}@caltech.edu} \\
}

\begin{document}

\maketitle
\vspace{-3mm}

\begin{abstract}
General-purpose foundation models for neural time series can help accelerate neuroscientific discoveries and enable applications such as brain computer interfaces (BCIs). A key component in scaling these models is population-level representation learning, which leverages information across channels to capture spatial as well as temporal structure. Population-level approaches have recently shown that such representations can be both efficient to learn on top of pretrained temporal encoders and produce useful representations for decoding a variety of downstream tasks. However, these models remain sensitive to mismatches in preprocessing, particularly on time-scales, between pretraining and downstream settings. We systematically examine how time-scale mismatches affects generalization and find that existing representations lack invariance. To address this, we introduce Time-scale Augmented Pretraining (TSAP), which consistently improves robustness to different time-scales across decoding tasks and builds invariance in the representation space. These results highlight handling preprocessing diversity as a key step toward building generalizable neural foundation models.

\end{abstract}

\section{Introduction} 

Building general-purpose representations of neural time series data is a foundational goal for neuroscience research. High-fidelity neural recordings such as intracranial electroencepholography (iEEG) capture complex activity patterns across multiple brain regions, but present significant modeling challenges due to inter-subject and session variability, and limited dataset sizes \citep{herff2020potential, jiang2025data}. As a result, many neuroscience studies and brain computer interface research use single-channel or subject-specific models, limiting their expressivity and generalizability \citep{willett2023high, wandelt2024representation, kunz2024representation, wang2024brain}.

Recent advances in self-supervised learning have inspired the development of foundation models for neural signals. Time-series foundation modeling and large-scale univariate pretraining provide robust and rich temporal embeddings for downstream decoding \citep{han2024capacity, ansari2024chronos, talukder2024totem, wang2023brainbert, liu2022seeing}. However, one major limitation is a lack of learned spatial information at the level of populations of channels. Population-level pretraining on top of these temporal embeddings is one promising approach to provide the missing spatial component, demonstrating strong performance on downstream tasks while being computationally and sample-efficient to train \citep{chau2025population, liu2023frequency}. Here we will focus on improving models that learn population-level representations from temporal encoders to enable efficient and generalizable scaling for neuroscience foundation models. 

\begin{figure}[t]
  \centering
  \includegraphics[width=\linewidth]{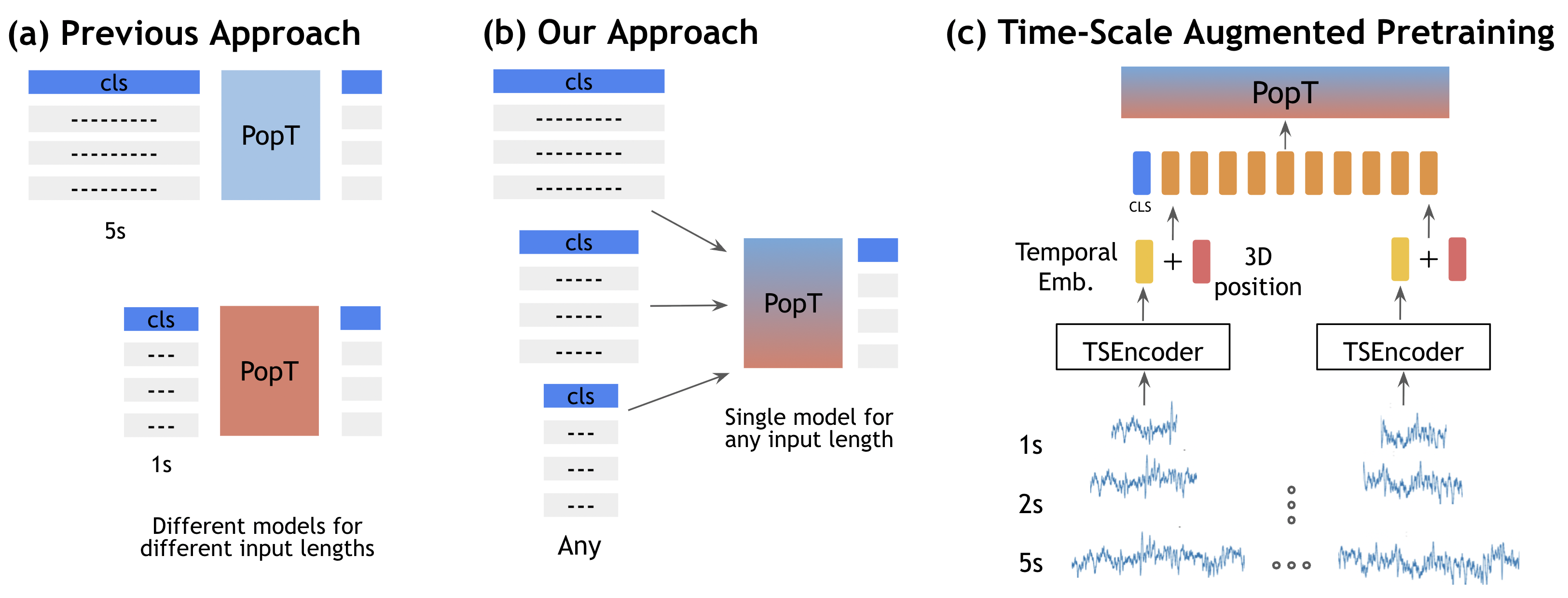}
  \caption{
  \textbf{Schematic of our approach}.
  (a) Previous population-level transformer (PopT) \citep{chau2025population} pretrain on fixed temporal windows, which opens up the question of how sensitive these models are to inputs of different time-scales. 
  (b) Our approach seeks to provide optimal performance for any input length. 
  (c) Time-scale Augmented Pretraining (TSAP) use samples from varying input interval lengths to achieve invariance to input time-scales. 
  }
  \label{approach}
\end{figure}

A key design limitation remains in this approach: the population-level layer is trained exclusively on the outputs of the temporal encoders. As a consequence, we empirically observe that these models are sensitive to preprocessing parameters used for input signals--such as the time-scales used for training. In practice, neural recordings vary widely in length across datasets and tasks \citep{peterson2022ajile12, wang2024brain, gwilliams2025hierarchical}, so building invariance into our models along this dimension is important. In this work, we quantify the degradation we get from ignoring these preprocessing mismatches, and explore techniques to remedy this. In particular, we propose a new strategy: Time-scale Augmented Pretraining (TSAP). Applied to iEEG decoding tasks, TSAP consistently improves generalization across seen and unseen time-scales, outperforming interval-specific baselines (\Cref{approach}b). Our analysis further demonstrates that TSAP reduces time-scale clustering in embedding space, leading to more invariant and transferable representations. Addressing this limitation is critical for realizing the promise of foundation models that are applicable across a wide range of tasks and experimental settings.

Our contributions are:
\begin{enumerate}
    \item A comparative evaluation of decoding performance on mismatched time-scales, confirming that pretraining and finetuning on the same interval length leads to better performance.
    \item A novel Time-scale Augmented Pretraining (TSAP) strategy that exposes the model to a spectrum of interval lengths, improving generalization, even to unseen lengths.
    \item Analysis of the embeddings from different time-scales across pretrained models. 
\end{enumerate}

\section{Methods}

\textbf{Population Transformer Framework.}
To investigate the time-scale invariance of spatial aggregation pretrained transformers, we adopt the Population Transformer (PopT) \citep{chau2025population} as our core framework, using architectural parameters and training configuration described in the original work. 

At a high level, for each electrode channel, $c$, a given interval $i$ of length $l$ (where $l$ can be varied) is pushed through a frozen temporal encoder, in this case we use BrainBERT \citep{wang2023brainbert}. These temporal embeddings are summed with a positional embedding derived from its 3D electrode coordinates (\Cref{approach}c), before being passed through the transformer encoder, yielding spatially contextualized channel representations and an aggregated \texttt{[CLS]} output. This \texttt{[CLS]} is then projected with a linear layer for downstream decoding. We adjust the PopT pretraining strategy below.

\textbf{Augmented Data and Training.} \label{TSAP_pretraining}To adapt the PopT framework for temporal invariance, we modified the data generation pipeline to expose the model to iEEG signals at multiple interval lengths. Specifically, we sampled recording segments of lengths $l \in {1, 2, 4, 5}$ seconds, with a fixed gap $g$ between consecutive windows for each input channel. Each interval was independently encoded into BrainBERT embeddings, yielding distinct temporal representations for the same underlying signal at different scales. These embeddings include overlapping windows across interval lengths, which the temporal encoder maps into non-identical representations. This augmentation strategy, TSAP, encourages the model to generalize across multiple time-scales.

\textbf{Embedding analysis.} To better understand the distributions of temporal embeddings and PopT representations, we perform 2D PCA on the embedding spaces. We take 100 samples from a specific subject-session from the Word Onset task, across the time-scales 1, 2, 3, 4, and 5 seconds. For temporal embedding PCA analysis, each sample is represented as a concatenation of all the channels embeddings, producing high-d $n_{chan} * h_{dim}$ vectors. For pretrained PopT \texttt{[CLS]} token analysis, each sample is pushed through pretrained 5s and TSAP PopT models, producing the processed \texttt{[CLS]} token representations. To analyze the clustering, we performed K-Means clustering with 5 means, aligned the clusters to the true classes based on the mode interval of points assigned to the cluster, and plotted the confusion matrices of cluster assignments to true intervals. 

\begin{figure}[b!]
  \centering
  \includegraphics[width=\linewidth]{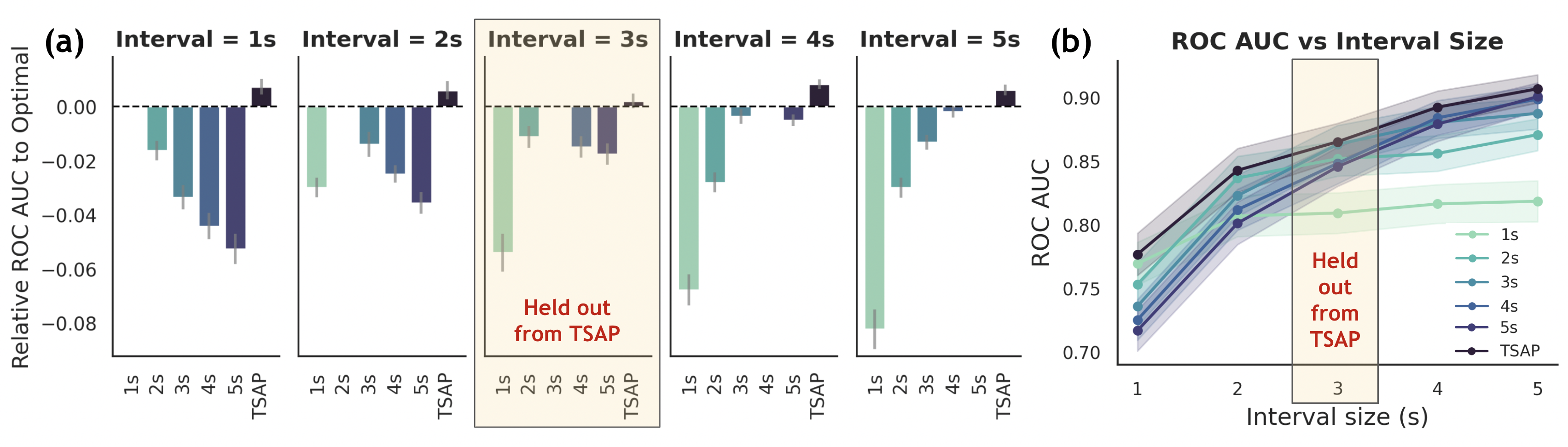}
  \caption{
  \textbf{Performance drop from mismatch in input time-scales is recovered by TSAP.} (a) Compared to the optimal (dotted line), models (x-axis) trained with mismatched time-scales perform much worse (below the line), while TSAP (dark blue) generally improves upon the optimal baseline. Shown are the Word Onset ROC AUC difference means and standard error across subjects and 5 seeds. (b) We see TSAP (dark blue) match or outperform all other models across all input lengths. Shown are the Word Onset ROC AUC mean and standard error across subjects and 5 seeds.
  }
  \label{results_performance}
\end{figure}

\section{Experiments}

\textbf{Data.}
We used iEEG, a type of neural time-series data collected via probes implanted within the brain to record local electrical signals at high temporal resolution and spatial precision. We used the publicly available BrainTreeBank dataset \citep{wang2024brain}. Data was collected from 10 subjects (total 1,688 electrodes, mean 167 electrodes per subject) who watched 26 movies (19 for pretraining and 7 for downstream decoding), while intracranial probes recorded their neural activity. We evaluate on two auditory-linguistic classification tasks from BrainTreeBank: (1) determining whether any speech at all is occurring (word onset), and (2) determining whether the beginning of a sentence is occurring (sentence onset). For each subject, we randomly select 90 electrodes to use for fine-tuning. 

\textbf{Baselines and Methods Compared.}
For our baselines, we compared against the non-pretrained PopT to assess the impact of pretraining, as well as the original PopT pretraining formulation as an optimal matched preprocessing baseline. To study the effect of interval-specific pretraining, we pretrained PopT on individual interval lengths of 1, 2, 3, 4, or 5 seconds to produce an "optimal" baseline model (5 seconds is the original PopT formulation). Each variant was trained following the same setup as the PopT paper, with the exception of using a learning rate of $1 \times 10^{-4}$ to improve training stability.  These experiments serve to test whether models pretrained and finetuned on the same interval length outperform non-corresponding counterparts across tasks and interval lengths.

\textbf{Evaluating TSAP.}
To understand how TSAP performs on a variety of downstream time-scales, we pretrain PopT with TSAP as described in \Cref{TSAP_pretraining} with interval lengths of 1, 2, 4, and 5 seconds (note that 3 seconds is held out). We doubled the number of pretraining steps from 500{,}000 to 1{,}000{,}000, while keeping the learning rate fixed at $1 \times 10^{-4}$ to allow the model to process the larger augmented dataset. For all pretraining runs, we selected the best-performing model checkpoint based on validation loss. After pretraining, we evaluated the TSAP model on all downstream interval lengths to compare its performance with the optimal baselines for held-in (1, 2, 4, and 5 seconds) and held-out (3 seconds) interval lengths. 

\textbf{Finetuning.}
For finetuning, embeddings were generated at interval lengths of 1, 2, 3, 4, or 5 seconds, with each sample labeled as positive or negative depending on whether it was centered around a word or sentence onset (depending on the task). Each finetuning experiment was performed on a single subject and interval length, with models evaluated across every subject-interval combination. The 3-second interval dataset served as the held-out interval set, as TSAP was not pretrained on 3-second interval data. To improve robustness, each experiment was repeated five times with different random seeds. For each seed, we selected the best model checkpoint based on validation ROC-AUC, and reported test performance on that model.

\begin{table*}[t!]
\small
\centering
\setlength{\tabcolsep}{3pt}
\begin{tabular}{@{}lc@{\hspace{1ex}}c@{\hspace{1.0ex}}c@{\hspace{1ex}}c@{\hspace{1.0ex}}c@{\hspace{1.0ex}}}
 & \textbf{1s Interval} & \textbf{2s Interval} & \textbf{3s Interval} & \textbf{4s Interval} & \textbf{5s Interval}\\\midrule
\textbf{Word Onset}: &  &  &  &  \\

\hspace{3mm} Non-Pretrained & $0.645 \pm 0.015$ & $0.665 \pm 0.027$ & $0.663 \pm 0.018$ & $0.671 \pm 0.019$ & $0.678 \pm 0.018$\\

\hspace{3mm} 1s Interval & $\underline{0.770} \pm \underline{0.017}$ & $0.807 \pm 0.016$ & $0.809 \pm 0.016$ & $0.817 \pm 0.016$ & $0.819 \pm 0.016$ \\

\hspace{3mm} 2s Interval & $0.753 \pm 0.017$ & $\underline{0.837} \pm \underline{0.017}$ & $0.852 \pm 0.014$ & $0.856 \pm 0.014$ & $0.871 \pm 0.013$ \\

\hspace{3mm} 3s Interval & $0.736 \pm 0.016$ & $0.823 \pm 0.016$ & $\underline{0.863} \pm \underline{0.015}$ & $0.881 \pm 0.013$ & $0.888 \pm 0.012$ \\

\hspace{3mm} 4s Interval & $0.725 \pm 0.016$ & $0.812 \pm 0.016$ & $0.849 \pm 0.016$ & $\underline{0.884} \pm \underline{0.014}$ & $0.899 \pm 0.010$ \\

\hspace{3mm} 5s Interval & $0.717 \pm 0.016$ & $0.801 \pm 0.017$ & $0.846 \pm 0.015$ & $0.879 \pm 0.014$ & $\underline{0.901} \pm \underline{0.011}$ \\

\hspace{3mm} \textbf{TSAP} & $\mathbf{ 0.777 } \pm \mathbf{ 0.017^* }$ & $\mathbf{ 0.843 } \pm \mathbf{ 0.017 }$ & $\mathbf{ 0.866 } \pm \mathbf{ 0.015 }$ & $\mathbf{ 0.893 } \pm \mathbf{ 0.013^*}$ & $\mathbf{ 0.907 } \pm \mathbf{ 0.011^*}$ 
\\\midrule
\textbf{Sentence Onset}: &&  &  &  \\
\hspace{3mm} Non-Pretrained & $0.715 \pm 0.021$ & $0.740 \pm 0.018$ & $0.731 \pm 0.018$ & $0.717 \pm 0.017$ & $0.710 \pm 0.018$\\

\hspace{3mm} 1s Interval & $\underline{0.790} \pm \underline{0.016}$ & $0.798 \pm 0.015$ & $0.785 \pm 0.015$ & $0.778 \pm 0.015$ & $0.776 \pm 0.014$ \\

\hspace{3mm} 2s Interval & $0.771 \pm 0.018$ & $\underline{0.837} \pm \underline{0.014}$ & $0.831 \pm 0.013$ & $0.828 \pm 0.014$ & $0.829 \pm 0.011$ \\

\hspace{3mm} 3s Interval & $0.764 \pm 0.016$ & $0.822 \pm 0.015$ & $\mathbf{0.846} \pm \mathbf{0.013}$ & $\underline{0.852} \pm \underline{0.012}$ & $0.853 \pm 0.011$ \\

\hspace{3mm} 4s Interval & $0.760 \pm 0.017$ & $0.803 \pm 0.016$ & $0.833 \pm 0.013$ & $0.851 \pm 0.013$ & $0.851 \pm 0.011$ \\

\hspace{3mm} 5s Interval & $0.760 \pm 0.016$ & $0.798 \pm 0.016$ & $0.821 \pm 0.014$ & $0.847 \pm 0.012$ & $\underline{0.860} \pm \underline{0.011}$ \\

\hspace{3mm} \textbf{TSAP} & $\mathbf{ 0.802 } \pm \mathbf{ 0.015^* }$ & $\mathbf{ 0.841 } \pm \mathbf{ 0.014 }$ & $\underline{0.843} \pm \underline{0.013}$ & $\mathbf{ 0.855 } \pm \mathbf{ 0.012 }$ & $\mathbf{ 0.865 } \pm \mathbf{ 0.010 }$ 
\\\midrule
\end{tabular}
\caption{
\textbf{Results across models and decoding tasks.} For each interval length (columns), we evaluate on models non-pretrained, pretrained with specific interval lengths (1s, 2s, 3s, 4s, and 5s), and pretrained on using TSAP (3s interval is heldout) (rows). We show two different downstream tasks: Word Onset and Sentence Onset (sections). Shown are the test ROC-AUC mean and standard error across subjects and 5 seeds per subject. Best per task is bolded, second best is underlined. Asterisks denote significant improvement compared to the optimal with paired t-test (\Cref{ttest_table_combined}).}
\label{performance_table}

\end{table*}

\section{Results}

\textbf{Decoding Performance.}
We sought to understand how much of a performance decrease we get from using mismatched input lengths between pretraining and finetuning. We find dramatic reductions in performance when these are mismatched (\Cref{results_performance}a). When we introduce TSAP, we recover and occasionally exceed the performance achieved by the optimal baseline models (\Cref{results_performance}). This is consistent across downstream time-scales, even for unseen time-scales such as the 3-second interval shown in \Cref{results_performance}b. We also see the same patterns across additional decoding tasks \Cref{performance_table}. By augmenting the pretraining task with additional time-scales, we obtain a model that is able to match or exceed the expected performance for each time-scale. 

\begin{figure}[h!]
  \centering
  \includegraphics[width=\linewidth]{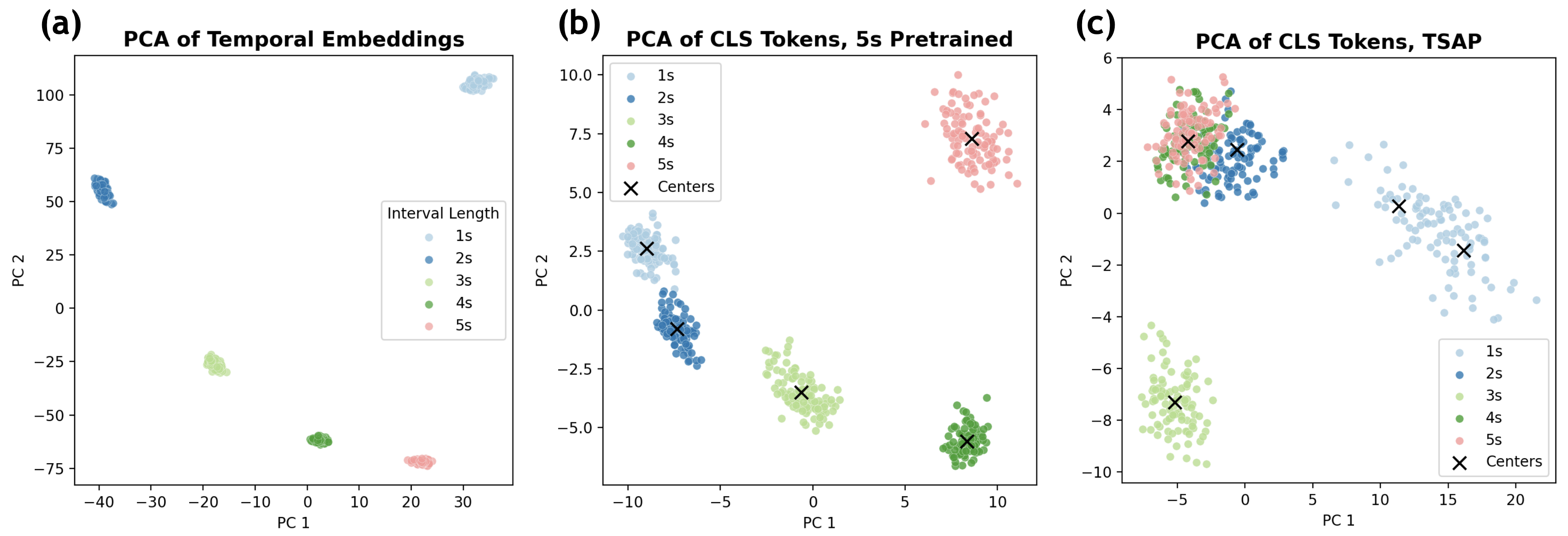}
  \caption{
  \textbf{PCA Analysis of Raw Embeddings and CLS tokens}. (a) PCA projection of temporal embeddings taken from different time-scales (colors) from 1 subject and 1 session from the Word Onset task. Temporal Embeddings tend to cluster by interval length despite them being from the same 100 samples. (b) PCA of CLS token after training with 5-second intervals only. We again see strong clustering by time-scale, with K-Means clusters identified for each ("X" marks). (c) PCA of CLS token after training with TSAP. We see that TSAP CLS tokens across several time-scales are clustered more closely together, with confused K-Means cluster ("X" marks).}
  \label{results_pca}
\end{figure}

\textbf{Embedding Space Analysis.}
We hypothesized that the temporal encoding representations of trials cropped to different time-scales would be represented drastically differently, despite containing overlapping information. To see if this is true, we project our BrainBERT encoded samples into the top 2 PCA component space, and find strong clustering by interval length (\Cref{results_pca}a). 

To check what was happening to the representations by doing TSAP versus training on only one length, we projected 100 samples with their \texttt{[CLS]} token representations from the two respective models. We find that the model trained only with 5-second time-scales still exhibits strong clustering of samples within their time-scales (\Cref{results_pca}b), while the TSAP model has much more overlap of its clusters suggesting greater time-scale invariance in its representations (\Cref{results_pca}c). The confusion matrices following K-means clustering of the \texttt{[CLS]} tokens further show how the 5-second Pretrained PopT produces extremely time-scale dependent clusters while there is more confusion with the cluster assignments with the TSAP model \Cref{results_confusion_matrix_appendix}.

\section{Discussion} 
Our exploration into the sensitivity of variable time-scale inputs reveals that current models are overfit to specific time-scale and preprocessing styles, preventing optimal performance across arbitrary time-scales (\Cref{results_performance}a). We find that pretraining with the same interval length provides optimal performance for evaluating on that time-scale. However, model performance drops as we evaluate on interval lengths different from the pretraining time-scale. Despite this, none of the pretrained models do as poorly as the non-pretrained performances, suggesting that there is still some valuable information being learned in pretraining despite mismatched time-scales (\Cref{performance_table}). To amend this sensitivity to input length, we tested an augmentation strategy that involves pretraining with samples of different interval lengths, and find that it is incredibly effective at recovering lost performance (\Cref{results_performance}) across all time-scale inputs. In fact, we find that our augmented pretraining strategy allows the resulting model to consistently outperform the optimal models in many cases, even performing better than most on held-out time-scale lengths. Models that can achieve optimal performance across time-scales can allow neuroscience and BCI research to use neural foundation models more effectively out of the box, providing optimal performance no matter the time-scale input size of the experimental paradigm. 

\textbf{Limitations and Future Work.}
We identify a gap in generalizability of current population-level foundation models to perform well across input time-scales, and explore a solution to remedy this. However, it remains to be seen how this solution compares with or augments other approaches, such as building invariance into the temporal encoders themselves \citep{zhang2022self, liu2023frequency, somaiya2022ts}, or leveraging smaller fixed patches and learning temporal and spatial components together \citep{wang2024cbramod, jiang2024large, zhang2024brant, talukder2024totem}. Future work could experiment with combinations of these approaches and evaluate which scales efficiently and provides best generalization to input time-scales.

\section{Conclusion}
Here we focused on improving population-level foundation models by addressing a key limitation in generalizability to time-scales. We quantify the drop in performance due to mismatches in preprocessing, and show that such performance drop can be easily recovered with our proposed TSAP technique. We further investigate the nature of the embeddings spaces and generalizability of this approach to unseen time-scales to demonstrate that it is an effective strategy for foundation models to adopt in order to gain generalizbility along this critical dimension.

\newpage

\bibliography{ref}
\bibliographystyle{plainnat}

\newpage

\appendix

\section{Related Works}
\label{related_works}

\textbf{Foundation Models for Neural Signals.}
Recent work on neural time-series—particularly EEG and iEEG—has leveraged foundation-model approaches to learn generalizable representations across individuals, tasks, and recording setups. Channel-independent pretraining has shown promise for spiking data \citep{liu2022seeing}, electrophysiological recordings \citep{wang2023brainbert, talukder2024totem, chau2024generalizability}, and general time-series \citep{talukder2024totem}, while related models have also been explored for EEG \citep{chien2022maeeg, kostas2021bendr, yi2023mmm}. However, these methods often focus on single channels or assume fixed sensor layouts, limiting their ability to capture population-level interactions across heterogeneous datasets. More recent efforts jointly pretrain spatial and temporal dimensions to handle variable inputs \citep{zhang2024brant, yang2024biot, jiang2024large, ye2024neural, cai2023mbrain}, but their coupled design increases computational cost and sensitivity to preprocessing. The Population Transformer (PopT) \citep{chau2025population} addresses this by combining pretrained temporal encoders with a spatial aggregation transformer to enable population-level modeling across electrode configurations, yet it remains constrained by the fixed-duration temporal inputs of its encoders, which can hinder transfer across mismatched time intervals and preprocessing schemes.

\textbf{General Time-Series Foundation Models and Variable-Length Training.}
Beyond the neural domain, several time-series approaches highlight the benefits of training with variable-length inputs. TS-Rep \citep{somaiya2022ts} encourages duration-agnostic representations via a triplet-based objective, while Time Warping with a Discriminative Teacher \citep{iwana2021time} introduces controlled distortions through dynamic time warping. Other methods exploit frequency-domain consistency, such as TF-C \citep{zhang2022self} and BioFAME \citep{liu2023frequency}, to promote invariance to temporal scale. These works demonstrate that temporal augmentation and multi-view learning can yield scale-robust features, but they are largely limited to univariate or single-modality settings. Additionally, many popular foundation models for time series \citep{ansari2024chronos, yue2022ts2vec, talukder2024totem, wang2023brainbert} still rely on fixed-length windows, requiring ad hoc strategies to handle context mismatches. This motivates our approach: augmenting PopT’s pretraining data with variable-length intervals to extend population-level neural representation learning toward temporal invariance in a multivariate setting.

\section{Training Details}
\label{training_details}
To run all our experiments (data processing, pretraining, evaluations, interpretability), one only needs 1 NVIDIA RTX A6000 (50GB GPU RAM). Pretraining PopT on a single interval takes approximately 1.5 days on 1 GPU and pretraining TSAP on a single interval takes approximately 3 days on 1 GPU. Our downstream evaluations take a few minutes to run each. For the purposes of data processing and gathering all the results in the paper, we parallelized the experiments on 6 GPUs.

\newpage
\section{Sentence Onset Results}
\label{sentence_onset_results}
\begin{figure}[h!]
  \centering
  \includegraphics[width=\linewidth]{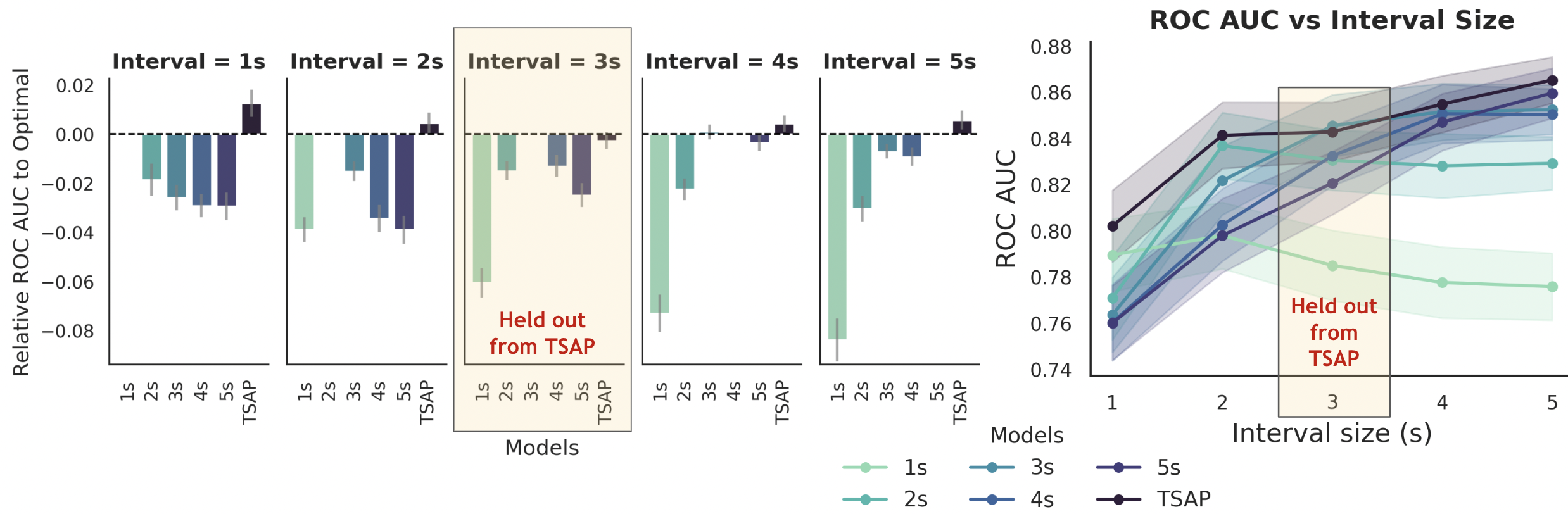}
  \caption{
  \textbf{Performance drop from mismatch in input time-scales is recovered by TSAP.} (a) Compared to the optimal (dotted line), models (x-axis) trained with mismatched time-scales perform much worse (below the line), while TSAP (dark blue) generally improves upon the optimal baseline. Shown are the Sentence Onset relative ROC AUC difference means and standard error across subjects and 5 seeds. (b) We see TSAP (dark blue) closely matches or outperform other models across all input time-scales. Shown are the Sentence Onset ROC AUC mean and standard error across subjects and 5 seeds.
  }
  \label{results_performance_appendix}
\end{figure}

\section{Cluster Analysis}
\label{cluster_analysis_appendix}
\begin{figure}[h!]
  \centering
  \includegraphics[width=\linewidth]{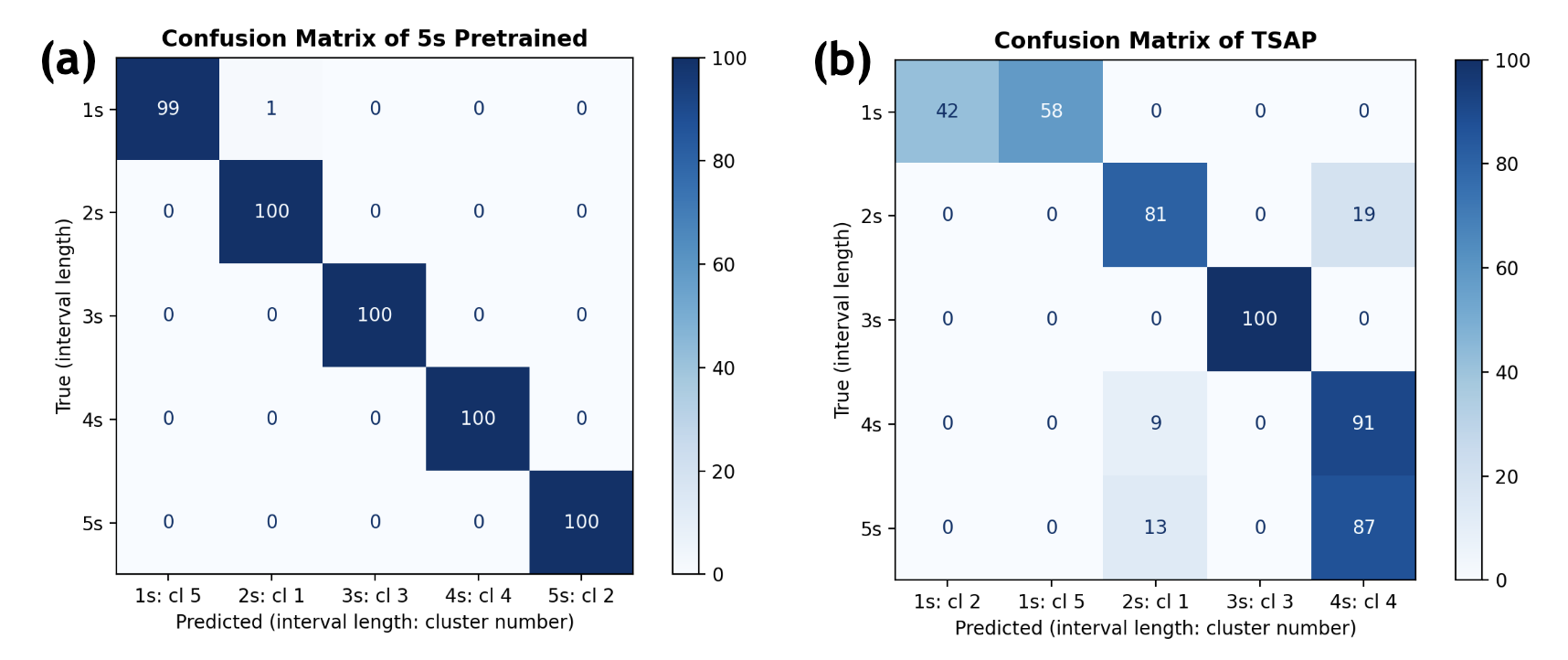}
  \caption{
  \textbf{Confusion matrix following K-Means clustering of CLS tokens.} For (a) 5s Pretrained PopT and (b) TSAP model. We see clean clustering in the 5s, but much more confusion in the TSAP version. 
  }
  \label{results_confusion_matrix_appendix}
\end{figure}

\newpage
\section{Statistical Analysis}

We chose to conduct a paired t-test due to the high variation between subjects and seeded splits. We see that most of the TSAP models are significantly better than the Optimal Baselines based on the p-values from this paired t-test. Additionally, the 95\% confidence interval have very small negative bounds which indicate high confidence that our model will perform at or better than the Optimal Baseline. The held-out 3-second interval has the least significant improvement for our TSAP model, but we still see that it occasionally has improvement upon the optimal baseline (1/2 wins). 
\begin{table*}[h!]
\small
\centering
\setlength{\tabcolsep}{6pt}
\begin{tabular}{@{}lcccccc@{}}
\textbf{Interval} & \textbf{Mean} & \textbf{Std. Err} & \textbf{$t$} & \textbf{$p$} & \textbf{95\% CI} & \textbf{$N$} \\
\midrule
\multicolumn{7}{l}{\textbf{Word Onset}}\\
\cmidrule(lr){1-7}
1s & $0.0074$ & $0.0029$ & $2.516$ & $0.01675*$ & $(0.0014,\,0.0134)$ & $35$ \\
2s & $0.0061$ & $0.0034$ & $1.774$ & $0.08506$ & $(-0.0009,\,0.0130)$ & $35$ \\
3s & $0.0021$ & $0.0027$ & $0.777$ & $0.4423$  & $(-0.0034,\,0.0076)$ & $35$ \\
4s & $0.0083$ & $0.0018$ & $4.647$ & $0.00005*$ & $(0.0047,\,0.0120)$ & $35$ \\
5s & $0.0062$ & $0.0020$ & $3.079$ & $0.00409*$ & $(0.0021,\,0.0102)$ & $35$ \\
\addlinespace
\multicolumn{7}{l}{\textbf{Sentence Onset}}\\
\cmidrule(lr){1-7}
1s & $0.0126$ & $0.0056$ & $2.258$ & $0.0305*$ & $(0.0013,\,0.0239)$ & $35$ \\
2s & $0.0045$ & $0.0042$ & $1.062$ & $0.2958$  & $(-0.0041,\,0.0131)$ & $35$ \\
3s & $-0.0026$ & $0.0034$ & $-0.782$ & $0.4396$ & $(-0.0095,\,0.0042)$ & $35$ \\
4s & $0.0042$ & $0.0033$ & $1.297$ & $0.2034$  & $(-0.0024,\,0.0109)$ & $35$ \\
5s & $0.0057$ & $0.0039$ & $1.463$ & $0.1526$  & $(-0.0022,\,0.0135)$ & $35$ \\
\bottomrule
\end{tabular}
\caption{
\textbf{Paired t-test results (TSAP vs Optimal Baseline)} for ROC AUC across subject/seed pairs. Shown are results for Word Onset (top) and Sentence Onset (bottom). Entries are the mean paired difference, standard error, $t$ statistic, $p$-value, 95\% confidence interval, and sample size ($N$).
}
\label{ttest_table_combined}
\end{table*}

\end{document}